# An Extended Čencov-Campbell Characterization of Conditional Information Geometry

**Guy Lebanon**
School of Computer Science
Carnegie Mellon University
lebanon@cs.cmu.edu

## Abstract

We formulate and prove an axiomatic characterization of conditional information geometry, for both the normalized and the non-normalized cases. This characterization extends the axiomatic derivation of the Fisher geometry by Čencov and Campbell to the cone of positive conditional models, and as a special case to the manifold of conditional distributions. Due to the close connection between the conditional $I$-divergence and the product Fisher information metric the characterization provides a new axiomatic interpretation of the primal problems underlying logistic regression and AdaBoost.

## 1 Introduction

The theory of information geometry presents a geometric interpretation of statistical properties and techniques. Among many examples of such properties and techniques are efficiency of estimators, robustness, maximum likelihood estimation for exponential models and hypothesis testing. The monograph of Amari and Nagaoka (2000) contains an overview of relevant research in this topic during the past fifty years.

A fundamental assumption in the information geometric framework, that manifests itself in many statistical and machine learning applications, is the choice of the Fisher information as the metric that underlies the geometry of probability distributions. The choice of the Fisher information metric may be motivated in several ways, the strongest of which is Čencov's characterization theorem (Lemma 11.3 in (Čencov, 1982)). In his theorem, Čencov proves that the Fisher information metric is the only metric that is invariant under a family of probabilistically meaningful mappings termed congruent embeddings by a markov morphism. Later on, Campbell (1986) stripped the proof from its category theory language and extended Čencov's result to include non-normalized models.

While the geometric approach to statistical inference has attracted considerable attention, little research was conducted on the geometric approach to conditional inference. The characterization theorems of Čencov and Campbell no longer apply in this setting and the different ways of choosing metrics on the manifold of conditional distributions, in contrast to the non-conditional case, are not supported by theoretical considerations.

In this paper we extend the results of Čencov and Campbell to provide an axiomatic characterization of conditional information geometry. We derive the characterization theorem in the setting of non-normalized conditional models from which the geometry for normalized models is obtained as a special case. In addition, we demonstrate a close connection between the characterized geometry and the conditional $I$ divergence which leads to a new axiomatic interpretation of the geometry underlying the primal problems of logistic regression and AdaBoost. This interpretation builds on the connection between AdaBoost and constrained minimization of $I$-divergence demonstrated by Lebanon and Lafferty (2001).

Throughout the paper we consider spaces of strictly positive conditional probability distributions where the sample spaces of the explanatory and response variable are finite. Moving to the infinite case presents some serious difficulties. The positivity constraint on the other hand does not play a crucial role in the theorem and may by discarded at some notational cost.

The next section describes some relevant concepts from Riemannian geometry and is followed by a description of the manifolds of normalized and non-normalized conditional models. Section 4 describes a family of probabilistic mappings that will serve as the basis for the invariance requirement of the characterization theorem in Section 5. Section 6 applies the characterization result to logistic regression and AdaBoost and is followed by concluding remarks.

## 2 Relevant Concepts from Riemannian Geometry

In this section we describe briefly relevant concepts from Riemannian geometry. For more details refer to any textbook discussing Riemannian geometry.

Informally, an $n$th dimensional differentiable manifold $\mathcal{M}$ is a continuous and differentiable set of points that is locally equivalent to $\mathbb{R}^n$. The set of infinitely differentiable real valued functions on $\mathcal{M}$ is denoted by $C^\infty(\mathcal{M})$. A diffeomorphism between two manifolds $\mathcal{M}, \mathcal{N}$ is a bijection $f : \mathcal{M} \to \mathcal{N}$ such that $f$ and $f^{-1}$ are infinitely differentiable.

With every point $p \in \mathcal{M}$, we associate a vector space $T_p\mathcal{M} \cong \mathbb{R}^n$ called the tangent space. The elements of $T_p\mathcal{M}$ can be thought of as directional derivatives at $p$ operating on $C^\infty(\mathcal{M})$. In many cases, the manifold $\mathcal{M}$ is a submanifold of a larger manifold, often $\mathbb{R}^m$, $m \geq n$. For example, the $n$th dimensional simplex

$$\mathbb{P}_n = \left\{ \theta \in \mathbb{R}^{n+1} : \quad \forall i \; \theta_i > 0, \quad \sum_{i=1}^{n+1} \theta_i = 1 \right\} \quad (1)$$

is a submanifold of $\mathbb{R}^{n+1}$. In these cases, the tangent space of the submanifold $T_p\mathcal{M}$ is a subspace of $T_p\mathbb{R}^m$ and we may represent tangent vectors $v \in T_p\mathcal{M}$ in the standard basis $\{\partial_i\}_{i=1}^m$ of the embedding tangent space $T_p\mathbb{R}^m$ as $v = \sum_{i=1}^m v_i \partial_i$.

A Riemannian manifold $(\mathcal{M}, g)$ is a differentiable manifold $\mathcal{M}$ equipped with a Riemannian metric $g$. The metric $g$ is defined by a symmetric positive definite inner product on the tangent spaces $T_p\mathcal{M}$ that is $C^\infty$ in $p$

$$g_p(\cdot, \cdot) : T_p\mathcal{M} \times T_p\mathcal{M} \to \mathbb{R}, \quad p \in \mathcal{M}.$$

Since for every $u, v \in T_p\mathcal{M}$

$$g_p(v, u) = \sum_{i=1}^n \sum_{j=1}^n v_i u_j g_p(\partial_i, \partial_j)$$

$g_p$ is completely described by $\{g_p(\partial_i, \partial_j)\}$ – the set of inner products between the basis elements $\{\partial_i\}_{i=1}^n$ of $T_p\mathcal{M}$.

The metric enables us to define lengths of tangent vectors $v \in T_p\mathcal{M}$ by $\|v\| = \sqrt{\langle v, v \rangle_p}$ and lengths of curves $\gamma : [a, b] \to \mathcal{M}$ by $L(\gamma) = \int_a^b \|\dot\gamma(t)\| dt$ where $\dot\gamma(t)$ is the velocity vector of the curve $\gamma$ at time $t$. Using the above definition of lengths of curves, we can define the distance $d(x, y)$ between two points $x, y \in \mathcal{M}$ as

$$d(x, y) = \inf_{\gamma \in \Gamma(x, y)} \int_a^b \|\dot\gamma(t)\| dt$$

where $\Gamma(x, y)$ is the set of piecewise differentiable curves connecting $x$ and $y$. The distance $d$, also called geodesic distance, turns the Riemannian manifold into a metric space. In other words the distance $d : \mathcal{M} \times \mathcal{M} \to \mathbb{R}$ defined above satisfies the usual properties of positivity, symmetry and triangle inequality.

Given two Riemannian manifolds $(\mathcal{M}, g)$, $(\mathcal{N}, h)$ and a diffeomorphism between them $f : \mathcal{M} \to \mathcal{N}$ we define the push-forward and pull-back maps below, which are of crucial importance to the characterization theorems of section 5.

**Definition 1.** *The push-forward map $f_* : T_x\mathcal{M} \to T_{f(x)}\mathcal{N}$, associated with $f : \mathcal{M} \to \mathcal{N}$ is defined by $f_*v$ obeying the relation*

$$v(r \circ f) = (f_*v)r, \quad \forall r \in C^\infty(\mathcal{N}).$$

Intuitively, the push-forward transforms velocity vectors of curves $\gamma$ to velocity vectors of transformed curves $f(\gamma)$.

**Definition 2.** *Given a metric $h$ on $\mathcal{N}$ we define a metric $f^*h$ on $\mathcal{M}$ called the pull-back metric by*

$$(f^*h)_x(u, v) = h_{f(x)}(f_*u, f_*v).$$

**Definition 3.** *An isometry is a diffeomorphism $f : \mathcal{M} \to \mathcal{N}$ between two Riemannian manifolds $(\mathcal{M}, g), (\mathcal{N}, h)$ for which the following condition holds*

$$g_x(u, v) = (f^*h)_x(u, v) \quad \forall x \in \mathcal{M}, \forall u, v \in T_x\mathcal{M}.$$

Isometries, as defined above, identify two Riemannian manifolds as identical in terms of their Riemannian structure. Accordingly, isometries preserve all the geometric properties including the geodesic distance function $d_g(x, y) = d_h(f(x), f(y))$. Note that the above definition of an isometry is defined through the local metric in contrast to the global definition of isometry in other branches of mathematical analysis.

## 3 Normalized and Non-Normalized Conditional Manifolds

Given a parametric family of distributions $\{p(y ; \theta) : \theta \in \Theta \subset \mathbb{R}^n\}$ and assuming some regularity conditions we can identify statistical models in it as points on the manifold $\Theta$. The Fisher information matrix $E_\theta\{ss^\top\}$ where $s = \nabla_\theta \log p(y ; \theta)$ may be used to endow $\Theta$ with a Riemannian structure

$$g_\theta(\partial_i, \partial_j) = E_\theta\{\partial_i \log p(y ; \theta) \; \partial_j \log p(y ; \theta)\}$$

where $\partial_i, \partial_j \in T_\theta\Theta$.

In the finite non-parametric setting, which is the topic of this paper, the sample space $\mathcal{Y}$ is a finite set with

$|\mathcal{Y}| = n$ and $\Theta$ is the $n-1$ simplex

$$\mathbb{P}_{n-1} = \left\{\theta \in \mathbb{R}^n : \forall i \; \theta_i > 0, \sum_{i=1}^n \theta_i = 1\right\}$$

which represents the manifold of all positive probability models over $\mathcal{Y}$. Representing tangent vectors $v \in T_\theta \mathbb{P}_{n-1}$ in the standard basis of $T_\theta \mathbb{R}^n$ as $v = \sum_{i=1}^n v_i \partial_i$ subject to the constraint $\sum_i v_i = 0$, the Fisher information metric becomes

$$g_\theta(u,v) = \sum_{i=1}^n \frac{u_i v_i}{\theta_i}, \quad u, v \in T_\theta \mathbb{P}_{n-1}.$$

For more details on the Fisher information metric in the parametric and the non-parametric cases refer to Amari and Nagaoka (2000).

Given two finite event sets $\mathcal{X}, \mathcal{Y}$ of sizes $k$ and $m$ respectively, a conditional probability model $p(y|x)$ reduces to an element of $\mathbb{P}_{m-1}$ for each $x \in \mathcal{X}$. We may thus identify the space of conditional probability models associated with $\mathcal{X}$ and $\mathcal{Y}$ as the product space

$$\mathbb{P}_{m-1} \times \cdots \times \mathbb{P}_{m-1} = \mathbb{P}_{m-1}^k.$$

For our purposes, it will be more convenient to work with the more general case of positive non-normalized conditional models. In this case the manifold of conditional models is the cone of $k \times m$ matrices with positive entries, denoted by $\mathbb{R}_+^{k \times m}$. Since $\mathbb{P}_{m-1}^k \subset \mathbb{R}_+^{k \times m}$, results obtained for non-normalized models apply to normalized models as a special case. In addition, some of the notation and formulation is simplified by working with non-normalized models. By taking this approach, we follow the philosophy of Campbell (1986) and Lebanon and Lafferty (2001).

We replace the cumbersome probabilistic notation with matrix notation $M_{ij} = p(y_j|x_i)$. Note that matrices that correspond to normalized models are (row) stochastic matrices. We denote tangent vectors to $\mathbb{R}_+^{k \times m}$ using the standard basis

$$T_M \mathbb{R}_+^{k \times m} = \text{span}\{\partial_{ij} : i = 1,\ldots,k, j = 1,\ldots,m\}.$$

Tangent vectors to $\mathbb{P}_{m-1}^k$, when expressed using the basis of the embedding tangent space $T_M \mathbb{R}_+^{k \times m}$ are linear combinations of $\{\partial_{ij}\}$ such that the sum of the combination coefficients over each row are 0, e.g.

$$\frac{1}{2}\partial_{11} + \frac{1}{2}\partial_{12} - \partial_{13} + \frac{1}{3}\partial_{21} - \frac{1}{3}\partial_{22} \in T_M \mathbb{P}_2^3.$$

Note that the identification of the space of conditional models as a product of simplexes demonstrates the topological and differentiable structure. In particular, we do not assume that the metric has a product form.

In the characterization theorem we will make use of the fact that $\mathbb{P}_{m-1}^k \cap \mathbb{Q}^{k \times m}$ and $\mathbb{R}_+^{k \times m} \cap \mathbb{Q}^{k \times m} = \mathbb{Q}_+^{k \times m}$ are dense in $\mathbb{P}_{m-1}^k$ and $\mathbb{R}_+^{k \times m}$ respectively. Since continuous functions are uniquely characterized by their values on dense sets, it is enough to compute the metric for positive rational models $\mathbb{Q}_+^{k \times m}$. The value of the metric on non-rational models follows from its continuous extension to $\mathbb{R}_+^{k \times m}$.

## 4 Congruent Embeddings by Markov Morphisms of Conditional Models

We now discuss some probabilistically meaningful transformations of conditional models that correspond to sufficient statistics on conditional models. These transformations play a crucial role in the characterization theorem and are termed congruent embeddings by a markov morphism for consistency with Čencov (1982) and Campbell (1986).

**Definition 4.** Let $M \in \mathbb{R}^{k \times m}$ and $Q = \{Q^{(i)}\}_{i=1}^k$ be a set of matrices in $\mathbb{R}^{m \times n}$. We define the row product $M \otimes Q \in \mathbb{R}^{k \times n}$ as

$$[M \otimes Q]_{ij} = \sum_{s=1}^m M_{is} Q_{sj}^{(i)} = [MQ^{(i)}]_{ij}. \qquad (2)$$

In other words, the $i$th row of $M \otimes Q$ is the $i$th row of the matrix product $MQ^{(i)}$.

**Definition 5.** Let $\mathcal{A} = \{A_1, \ldots, A_m\}$ be a partition of $\{1, \ldots, n\}$ with $0 < m \leq n$. An $\mathcal{A}$-stochastic matrix is a stochastic matrix $Q \in \mathbb{R}^{m \times n}$ whose rows are concentrated on $\mathcal{A}$. In other words, $\forall i \; \sum_{j=1}^n Q_{ij} = 1$ and

$$Q_{ij} = \begin{cases} c_{ij} > 0 & j \in A_i \\ 0 & j \notin A_i \end{cases}.$$

For example, if $\mathcal{A} = \{\{1,3\},\{2,4\},\{5\}\}$ then the following matrix is $\mathcal{A}$-stochastic

$$\begin{pmatrix} 1/3 & 0 & 2/3 & 0 & 0 \\ 0 & 1/2 & 0 & 1/2 & 0 \\ 0 & 0 & 0 & 0 & 1 \end{pmatrix}.$$

Note that the columns of any $\mathcal{A}$-stochastic matrix have precisely one non-zero element.

**Definition 6.** An $\mathcal{A}$-stochastic matrix is called uniform if every row has the same number of non-zero elements and if all its positive entries are identical.

For example, the following matrix is a uniform $\mathcal{A}$-stochastic matrix for $\mathcal{A} = \{\{1,3\},\{2,4\},\{5,6\}\}$

$$\begin{pmatrix} 1/2 & 0 & 1/2 & 0 & 0 & 0 \\ 0 & 1/2 & 0 & 1/2 & 0 & 0 \\ 0 & 0 & 0 & 0 & 1/2 & 1/2 \end{pmatrix}.$$

The definition below of a congruent embedding by a markov morphism is a straightforward generalization of Čencov and Campbell's definition to the case of conditional models.

**Definition 7.** *Let $\mathcal{B}$ be a $k$ sized partition of $\{1,\ldots,l\}$ and $\{\mathcal{A}^{(i)}\}_{i=1}^{k}$ be a set of $m$ sized partitions of $\{1,\ldots,n\}$. For a $\mathcal{B}$-stochastic matrix $R \in \mathbb{R}_+^{k\times l}$ and $Q = \{Q^{(i)}\}_{i=1}^{k}$ a sequence of $\mathcal{A}^{(i)}$-stochastic matrices in $\mathbb{R}_+^{m\times n}$, we define the injection $f:\mathbb{R}_+^{k\times m}\to \mathbb{R}_+^{l\times n}$*

$$f(M) = R^\top(M\otimes Q). \tag{3}$$

*Maps of the form (3) are termed congruent embeddings by a markov morphism of $\mathbb{R}_+^{k\times m}$ into $\mathbb{R}_+^{l\times n}$ and are denoted by the function class $\mathfrak{F}_{k,m}^{l,n}$.*

The componentwise version of equation (3) is

$$[f(M)]_{ij} = \sum_{s=1}^{k}\sum_{t=1}^{m} R_{si}Q_{tj}^{(s)}M_{st} \tag{4}$$

with the above sum containing precisely one non-zero term since every column of $Q^{(s)}$ and $R$ contains only one non-zero entry. The push-forward map $f_*: T_M\mathbb{R}_+^{k\times m} \to T_{f(M)}\mathbb{R}_+^{l\times n}$ associated with $f$ is

$$f_*(\partial_{ab}) = \sum_{i=1}^{l}\sum_{j=1}^{n} R_{ai}Q_{bj}^{(a)}\partial'_{ij} \tag{5}$$

for $\partial_{ab} \in T_M\mathbb{R}_+^{k\times m}$ and $\partial'_{ij} \in T_{f(M)}\mathbb{R}_+^{l\times n}$. Congruent embeddings by a markov morphism are injective and if restricted to the space of normalized models $\mathbb{P}_{m-1}^k$ we have $f(\mathbb{P}_{m-1}^k) \subset \mathbb{P}_{n-1}^l$.

Using equation (5), the pull-back of a metric $g$ on $\mathbb{R}_+^{l\times n}$ through $f\in\mathfrak{F}_{k,m}^{l,n}$ is

$$(f^*g)_M(\partial_{ab},\partial_{cd}) = g_{f(M)}(f_*\partial_{ab}, f_*\partial_{cd}) \tag{6}$$
$$= \sum_{i=1}^{l}\sum_{j=1}^{n}\sum_{s=1}^{l}\sum_{t=1}^{n} R_{ai}R_{cs}Q_{bj}^{(a)}Q_{dt}^{(c)} g_{f(M)}(\partial'_{ij},\partial'_{st})$$

for $\partial_{ab},\partial_{cd} \in T_M\mathbb{R}_+^{k\times m}$.

The term congruent embedding by a markov morphism is motivated by the following observation. A map $f\in\mathfrak{F}_{k,m}^{l,n}$ transforms $p$, a conditional model over explanatory space $\{x_1,\ldots,x_k\}$ and response space $\{y_1,\ldots,y_m\}$, to a conditional model $q = f(p)$ over explanatory space $\{x_1,\ldots,x_l\}$ and response space $\{y_1,\ldots,y_n\}$ in the following way. If $R = I$ the model $p(\cdot|x_i)$ corresponds to extracting a sufficient statistic represented by the rows of $Q^{(i)}$ from the model $q(\cdot|x_i)$. If $R \neq I$, then in addition a sufficient statistic on the explanatory space $\{x_1,\ldots,x_l\}$ is extracted to obtain $\{x_1,\ldots,x_k\}$.

Several observations are in order. First, note that the above definitions of the manifold of conditional models and of $\mathfrak{F}_{k,m}^{l,n}$ are applicable in a non-parametric setting. In this setting, statistics correspond to taking the row product of a model with a stochastic matrix. Sufficient statistics correspond to taking the row product of the model with an $\mathcal{A}$-stochastic matrix, that contains only one non-zero element in each column. In this case the mapping is injective and hence invertible.

## 5 A Čencov Characterization of Metrics on Conditional Manifolds

We now turn to the generalization of Čencov's theorem to conditional manifolds. We proceed with the case of non-normalized conditional models and come back at the end of the section to normalized models. The method of proof is somewhat similar to the generalization of Čencov's theorem to the positive cone by Campbell (1986). We use only standard techniques from differential geometry and avoid the use of category theory. Before we state the main theorems we present some results and definitions that will be useful in the proofs.

We denote by $M_i$ the $i$th row of the matrix $M$ and by $|\cdot|$ the $L^1$ norm applied to vectors or matrices. For non-negative matrices $M$ we have $|M| = \sum_i|M_i| = \sum_{ij} M_{ij}$.

**Proposition 1.** *Maps in $\mathfrak{F}_{k,m}^{l,n}$ are norm preserving:*

$$|M| = |f(M)| \qquad \forall f \in \mathfrak{F}_{k,m}^{l,n},\ \forall M \in \mathbb{R}_+^{k\times m}.$$

*Proof.* Multiplying a positive row vector $v$ by an $\mathcal{A}$-stochastic matrix $T$ is norm preserving

$$|vT| = \sum_i [vT]_i = \sum_j v_j \sum_i T_{ji} = |v|.$$

As a result, $|[MQ^{(i)}]_i| = |M_i|$ for any positive matrix $M$ and hence $|M\otimes Q| = |M|$. A map $f\in\mathfrak{F}_{k,m}^{l,n}$ is norm preserving since

$$|M| = |M\otimes Q| = |(M\otimes Q)^\top| = |(M\otimes Q)^\top R|$$
$$= |R^\top(M\otimes Q)| = |f(M)|.$$

□

We discuss next three mappings in $\mathfrak{F}_{k,m}^{l,n}$ that will have a key role in the proof of the characterization theorem.

### 5.1 Three Useful Transformations

The first transformation $\mathfrak{h}_\sigma^\Pi \in \mathfrak{F}_{k,m}^{k,m}$ is parameterized by a permutation $\sigma$ over $k$ letters and a sequence of permutations $\Pi = (\pi^{(1)}, \ldots, \pi^{(k)})$ over $m$ letters. It is defined by $Q^{(i)}$ being the permutation matrix that corresponds to $\pi^{(i)}$ and $R$ being the permutation matrix that corresponds to $\sigma$. The push forward is $\mathfrak{h}_{\sigma*}^\Pi(\partial_{ab}) = \partial'_{\sigma(a)\pi^{(a)}(b)}$ and requiring $\mathfrak{h}_\sigma^\Pi$ to be an isometry from $(\mathbb{R}_+^{k\times m}, g)$ to itself results in

$$g_M(\partial_{ab}, \partial_{cd}) = g_{\mathfrak{h}_\sigma^\Pi(M)}(\partial_{\sigma(a)\pi^{(a)}(b)}, \partial_{\sigma(c)\pi^{(c)}(d)}). \quad (7)$$

The usefulness of $\mathfrak{h}_\sigma^\Pi$ stems in part from the following proposition.

**Proposition 2.** *Given $\partial_{a_1b_1}, \partial_{a_2b_2}, \partial_{c_1d_1}, \partial_{c_2d_2}$ with $a_1 \neq c_1$ and $a_2 \neq c_2$ there exists $\sigma, \Pi$ such that*

$$\mathfrak{h}_{\sigma*}^\Pi(\partial_{a_1b_1}) = \partial_{a_2b_2} \qquad \mathfrak{h}_{\sigma*}^\Pi(\partial_{c_1d_1}) = \partial_{c_2d_2}. \quad (8)$$

*Proof.* The desired map may be obtained by selecting $\Pi, \sigma$ such that $\sigma(a_1) = a_2, \sigma(c_1) = c_2$ and $\pi^{(a_1)}(b_1) = b_2$, $\pi^{(c_1)}(d_1) = d_2$. □

The second transformation $\mathfrak{r}_{zw} \in \mathfrak{F}_{k,m}^{kz,mw}$ for $z, w \in \mathbb{N}$, is defined by $Q^{(1)} = \cdots = Q^{(k)} \in \mathbb{R}^{m \times mw}$ and $R \in \mathbb{R}^{k \times kz}$ being uniform matrices. The exact forms of $\{Q^{(i)}\}$ and $R$ are immaterial for our purposes. By equation (5) the push-forward is

$$\mathfrak{r}_{zw*}(\partial_{st}) = \frac{1}{zw} \sum_{i=1}^{z} \sum_{j=1}^{w} \partial'_{\pi(i)\sigma(j)}$$

for some permutations $\pi, \sigma$ that depend on $s, t$. The pull-back of $g$ is

$$(\mathfrak{r}_{zw}^*g)_M(\partial_{ab}, \partial_{cd}) = \quad (9)$$
$$\frac{1}{(zw)^2} \sum_{i=1}^{z} \sum_{j=1}^{w} \sum_{s=1}^{z} \sum_{t=1}^{w} g_{\mathfrak{r}_{zw}(M)}(\partial'_{\pi_1(i),\sigma_1(j)}, \partial'_{\pi_2(s),\sigma_2(t)}).$$

The third transformation $\mathfrak{h}_M$ is parameterized by a rational model $M \in \mathbb{Q}_+^{k \times m}$. Since $M$ is rational it may be written as

$$M = \frac{1}{z}\tilde{M}, \qquad \tilde{M} \in \mathbb{N}^{k \times m} \ z \in \mathbb{N}$$

where $\mathbb{N}$ is the natural numbers. The mapping $\mathfrak{h}_M \in \mathfrak{F}_{k,m}^{|\tilde{M}|, \prod_s |\tilde{M}_s|}$ is associated with $Q^{(i)} \in \mathbb{R}^{m \times \prod_s |\tilde{M}_s|}$ and $R \in \mathbb{R}^{k \times |\tilde{M}|}$ which are defined as follows. The $j$th row of $Q^{(i)}$ is required to have $\tilde{M}_{ij} \prod_{s \neq i} |\tilde{M}_s|$ non-zero elements of value $(\tilde{M}_{ij} \prod_{s \neq i} |M_s|)^{-1}$ while the $i$th row of $R$ is required to have $|\tilde{M}_i|$ non-zero elements of value $|\tilde{M}_i|^{-1}$. Note that the matrices $\{Q^{(i)}\}$ have the same number of columns since $\sum_j \tilde{M}_{ij} \prod_{s \neq i} |\tilde{M}_s| = \prod_s |\tilde{M}_s|$ which is independent of $i$. The exact forms of $\{Q^{(i)}\}$ and $R$ do not matter for our purposes as long as the above restrictions and the requirements of definition 7 apply.

The usefulness of $\mathfrak{h}_M$ comes from the fact that it transforms $M$ into a constant matrix.

**Proposition 3.** *For $M = \frac{1}{z}\tilde{M} \in \mathbb{Q}_+^{k \times m}$,*

$$\mathfrak{h}_M(M) = \left(z \prod_s |\tilde{M}_s|\right)^{-1} \mathbf{1}$$

*where $\mathbf{1}$ is a matrix of ones of size $|\tilde{M}| \times \prod_s |\tilde{M}_s|$.*

*Proof.* $[M \otimes Q]_i$ is a row vector of size $\prod_s |\tilde{M}_s|$ whose elements are $(z \prod_{s \neq i} |\tilde{M}_s|)^{-1}$. Multiplying on the left by $R$ results in $[R(M \otimes Q)]_{ij} = (z \prod_s |\tilde{M}_s|)^{-1}$. □

By equation (5) the push-forward of $\mathfrak{h}_M$ is

$$\mathfrak{h}_{M*}(\partial_{st}) = \frac{\sum_{i=1}^{|\tilde{M}_s|} \sum_{j=1}^{\tilde{M}_{st} \prod_{l \neq s} |\tilde{M}_l|} \partial'_{\pi(i)\sigma(j)}}{\tilde{M}_{st} \prod_i |\tilde{M}_i|}. \quad (10)$$

for some permutations $\pi, \sigma$ that depend on $M, s, t$. The pull-back is

$$(\mathfrak{h}_M^* g)_M(\partial_{ab}, \partial_{cd}) =$$
$$\frac{\sum_i \sum_s \sum_j \sum_t g_{\mathfrak{h}_M(M)}(\partial_{\pi_1(i)\sigma_1(j)}, \partial_{\pi_2(s)\sigma_2(t)})}{\tilde{M}_{ab}\tilde{M}_{cd} \prod_s |\tilde{M}_s|^2} \quad (11)$$

where the first two summations are over $1, \ldots, |\tilde{M}_a|$ and $1, \ldots, |\tilde{M}_c|$ and the last two summations are over $1, \ldots, \tilde{M}_{ab} \prod_{l \neq a} |\tilde{M}_l|$ and $1, \ldots, \tilde{M}_{cd} \prod_{l \neq c} |\tilde{M}_l|$.

### 5.2 The Characterization Theorem

Theorems 1 and 2 below are the main result of the paper.

**Theorem 1.** *Let $\{(\mathbb{R}_+^{k \times m}, g^{(k,m)}) : k \geq 1, m \geq 2\}$ be a sequence of Riemannian manifolds with the property that every congruent embedding by a markov morphism is an isometry. Then*

$$g_M^{(k,m)}(\partial_{ab}, \partial_{cd}) =$$
$$A(|M|) + \delta_{ac}\left(\frac{|M|}{|M_a|}B(|M|) + \delta_{bd}\frac{|M|}{M_{ab}}C(|M|)\right) \quad (12)$$

*for some $A, B, C \in C^\infty(\mathbb{R}_+)$.*

*Proof.* The proof below uses the isometry requirement to obtain restrictions on $g_M^{(k,m)}(\partial_{ab}, \partial_{cd})$ first for $a \neq c$, followed by the case of $a = c, b \neq d$ and finally for the case $a = c, b = d$. In each of these cases, we first characterize the metric at constant matrices $U$ and then compute it for rational models $M$ by pulling back the metric at $U$ through $\mathfrak{y}_M$. The value of the metric at non-rational models follows from the rational case by the denseness of $\mathbb{Q}_+^{k \times m}$ in $\mathbb{R}_+^{k \times m}$ and the continuity of the metric.

*Part I:* $g_M^{(k,m)}(\partial_{ab}, \partial_{cd})$ *for* $a \neq c$

We start by computing the metric at constant matrices $U$. Given $\partial_{a_1 b_1}, \partial_{c_1 d_1}, a_1 \neq c_1$ and $\partial_{a_2 b_2}, \partial_{c_2 d_2}, a_2 \neq c_2$ we can use Proposition 2 and equation (7) to pull back through a corresponding $\mathfrak{h}_\sigma^\Pi$ to obtain

$$g_U^{(k,m)}(\partial_{a_1 b_1}, \partial_{c_1 d_1}) = g_U^{(k,m)}(\partial_{a_2 b_2}, \partial_{c_2 d_2}).$$

As a result we have that for $a \neq c$, $g_U^{(k,m)}(\partial_{ab}, \partial_{cd})$ depends only on $k, m$ and $|U|$ and we denote it temporarily by $\hat{A}(k, m, |U|)$.

Pulling back $g^{(kz,mw)}$ through $\mathfrak{r}_{zw}$ according to equation (9) we have

$$\hat{A}(k, m, \alpha) = g_U^{(k,m)}(\partial_{ab}, \partial_{cd})$$
$$= \frac{(zw)^2}{(zw)^2} \hat{A}(kz, mw, |\mathfrak{r}_{zw}(U)|) = \hat{A}(kz, mw, |U|) \quad (13)$$

since $\mathfrak{r}_{zw}(U)$ is a constant matrix with the same norm as $U$. Equation (13) holds for any $z, w \in \mathbb{N}$ and hence $g_U^{(k,m)}(\partial_{ab}, \partial_{cd})$ does not depend on $k, m$ and we write

$$g_U^{(k,m)}(\partial_{ab}, \partial_{cd}) = A(|U|) \quad \text{for some} \quad A \in C^\infty(\mathbb{R}_+).$$

We turn now to computing $g_M^{(k,m)}(\partial_{ab}, \partial_{cd}), a \neq c$ for rational models $M = \frac{1}{z}\tilde{M}$. Pulling back through $\mathfrak{y}_M$ according to equation (11) we have

$$g_M^{(k,m)}(\partial_{ab}, \partial_{cd}) = \frac{\tilde{M}_{ab} \tilde{M}_{cd} \prod_s |\tilde{M}_s|^2}{\tilde{M}_{ab} \tilde{M}_{cd} \prod_s |\tilde{M}_s|^2} A(|\mathfrak{y}_M(M)|)$$
$$= A(|M|). \quad (14)$$

Finally, since $\mathbb{Q}_+^{k \times m}$ is dense in $\mathbb{R}_+^{k \times m}$ and $g_M^{(k,m)}$ is continuous in $M$, equation (14) holds for all models in $\mathbb{R}_+^{k \times m}$.

*Part II:* $g_M^{(k,m)}(\partial_{ab}, \partial_{cd})$ *for* $a = c, b \neq d$

As before we start with constant matrices $U$. Given $\partial_{a_1 b_1}, \partial_{c_1 d_1}$ with $a_1 = c_1, b_1 \neq d_1$ and $\partial_{a_2 b_2}, \partial_{c_2 d_2}$ with $a_2 = c_2, b_2 \neq d_2$ we can pull-back through $\mathfrak{h}_\sigma^\Pi$ with $\sigma(a_1) = a_2, \pi^{(a_1)}(b_1) = b_2$ and $\pi^{(a_1)}(d_1) = d_2$ to obtain

$$g_U^{(k,m)}(\partial_{a_1 b_1}, \partial_{c_1 d_1}) = g_U^{(k,m)}(\partial_{a_2 b_2}, \partial_{c_2 d_2}).$$

It follows that $g_U^{(k,m)}(\partial_{ab}, \partial_{ad})$ depends only on $k, m, |U|$ and we temporarily denote

$$g_U^{(k,m)}(\partial_{ab}, \partial_{ad}) = \hat{B}(k, m, |U|).$$

Pulling back through $\mathfrak{r}_{zw}$ (9) we obtain

$$\hat{B}(k, m, \alpha) = \frac{zw^2 \hat{B}(kz, mw, |U|)}{(zw)^2} + \frac{z(z-1)w^2 A(|U|)}{(zw)^2}$$
$$= \frac{1}{z} \hat{B}(kz, mw, |U|) + \frac{z-1}{z} A(|U|).$$

Rearranging and dividing by $k$ results in

$$\frac{\hat{B}(k, m, |U|) - A(|U|)}{k} = \frac{\hat{B}(kz, mw, |U|) - A(|U|)}{kz}.$$

It follows that $\frac{\hat{B}(k,m,|U|) - A(|U|)}{k} = B(|U|)$ for some $B \in C^\infty(\mathbb{R}_+)$ and we may write

$$g_U^{(k,m)}(\partial_{ab}, \partial_{ad}) = A(|U|) + kB(|U|). \quad (15)$$

Next, we compute the metric for positive rational matrices $M = \frac{1}{z}\tilde{M}$. Pulling back through $\mathfrak{y}_M$ (11) we obtain

$$g_M^{(k,m)}(\partial_{ab}, \partial_{ad}) \quad (16)$$
$$= \frac{|\tilde{M}_a| - 1}{|\tilde{M}_a|} A(|M|) + \frac{1}{|\tilde{M}_a|} \Big( A(|M|) + B(|M|) \sum_i |\tilde{M}_i| \Big)$$
$$= A(|M|) + \frac{|\tilde{M}|}{|\tilde{M}_a|} B(|M|) = A(|M|) + \frac{|M|}{|M_a|} B(|M|).$$

As previously, by denseness of $\mathbb{Q}_+^{k \times m}$ in $\mathbb{R}_+^{k \times m}$ and continuity of $g^{(k,m)}$ equation (16) holds for all $M \in \mathbb{R}_+^{k \times m}$.

*Part III:* $g_M^{(k,m)}(\partial_{ab}, \partial_{cd})$ *for* $a = c, b = d$

As before, we start by computing the metric for constant matrices $U$. Given $a_1, b_1, a_2, b_2$ we pull back through $\mathfrak{h}_\sigma^\Pi$ with $\sigma(a_1) = a_2, \pi^{(a_1)}(b_1) = b_2$ to obtain

$$g_U^{(k,m)}(\partial_{a_1 b_1}, \partial_{a_1 b_1}) = g_U^{(k,m)}(\partial_{a_2 b_2}, \partial_{a_2 b_2}).$$

It follows that $g_U^{(k,m)}(\partial_{ab}, \partial_{ab})$ does not depend on $a, b$ and we temporarily denote

$$g_U^{(k,m)}(\partial_{ab}, \partial_{ab}) = \hat{C}(k, m, |U|).$$

Pulling back through $\mathfrak{r}_{zw}$ (9) we obtain

$$\hat{C}(k, m, |U|) = \frac{zw \hat{C}(kz, mw, |U|)}{(zw)^2} + \frac{z(z-1)w^2 A(|U|)}{(zw)^2}$$
$$+ \frac{zw(w-1)(A(|U|) + kzB(|U|))}{(zw)^2} = \frac{\hat{C}(kz, mw, |U|)}{zw}$$
$$+ \Big(1 - \frac{1}{zw}\Big) A(|U|) + \Big(k - \frac{zk}{zw}\Big) B(|U|)$$

which after rearrangement and dividing by $km$ is

$$\frac{\hat{C}(k, m, |U|) - A(|U|) - kB(|U|)}{km} =$$
$$\frac{\hat{C}(kz, mw, |U|) - A(|U|) - kzB(|U|)}{kzmw}. \quad (17)$$

It follows that the left side of (17) equals a function $C(|U|)$ for some $C \in C^\infty(\mathbb{R}_+)$ independent of $k$ and $m$ resulting in

$$g_U^{(k,m)}(\partial_{ab}, \partial_{ab}) = A(|U|) + kB(|U|) + kmC(|U|).$$

Finally, we consider $g_M^{(k,m)}(\partial_{ab}, \partial_{ab})$ for positive rational matrices $M = \frac{1}{z}\tilde{M}$. Pulling back through $\mathfrak{y}_M$ (11)

$$g_M^{(k,m)}(\partial_{ab}, \partial_{ab}) = \frac{|\tilde{M}_a| - 1}{|\tilde{M}_a|} A(|M|)$$
$$+ \left(\frac{1}{|\tilde{M}_a|} - \frac{1}{\tilde{M}_{ab} \prod_i |\tilde{M}_i|}\right) \left(A(|M|) + B(|M|) \sum_i |\tilde{M}_i|\right)$$
$$+ \frac{A(|M|) + B(|M|) \sum_i |\tilde{M}_i| + C(|M|) \prod_j |\tilde{M}_j| \sum_i |\tilde{M}_i|}{\tilde{M}_{ab} \prod_i |\tilde{M}_i|}$$
$$= A(|M|) + \frac{|\tilde{M}|}{|\tilde{M}_a|} B(|M|) + \frac{|\tilde{M}|}{\tilde{M}_{ab}} C(|M|)$$
$$= A(|M|) + \frac{|M|}{|M_a|} B(|M|) + \frac{|M|}{M_{ab}} C(|M|). \quad (18)$$

Since the positive rational matrices are dense in $\mathbb{R}_+^{k \times m}$ and the metric $g_M^{(k,m)}$ is continuous in $M$, equation (18) holds for all models $M \in \mathbb{R}_+^{k \times m}$ completing the last stage of the proof. □

The following theorem is the converse of Theorem 1.

**Theorem 2.** *Let $\{(\mathbb{R}_+^{k \times m}, g^{(k,m)})\}$ be a sequence of Riemannian manifolds, with the metrics $g^{(k,m)}$ given by equation (12). Then every congruent embedding by a markov morphism is an isometry.*

*Proof.* Considering arbitrary $M \in \mathbb{R}_+^{k \times m}$ and $f \in \mathfrak{F}_{k,m}^{l,n}$ we have by equation (6)

$$g_{f(M)}^{(l,n)}(f_*\partial_{ab}, f_*\partial_{cd}) \quad (19)$$
$$= \sum_{i=1}^l \sum_{j=1}^n \sum_{s=1}^l \sum_{t=1}^n R_{ai} R_{cs} Q_{bj}^{(a)} Q_{dt}^{(c)} g_{f(M)}^{(l,n)}(\partial'_{ij}, \partial'_{st}).$$

For $a \neq c$ equation (19) reduces to

$$A(|f(M)|) \sum_{i=1}^l \sum_{j=1}^n \sum_{s=1}^l \sum_{t=1}^n R_{ai} R_{cs} Q_{bj}^{(a)} Q_{dt}^{(c)}$$
$$= A(|f(M)|) = A(|M|) = g_M^{(k,m)}(\partial_{ab}, \partial_{cd})$$

since $R$ and $Q^{(i)}$ are stochastic matrices.

For $a = c, b \neq d$, equation (19) reduces to

$$A(|f(M)|) \sum_{i=1}^l \sum_{j=1}^n \sum_{s=1}^l \sum_{t=1}^n R_{ai} R_{cs} Q_{bj}^{(a)} Q_{dt}^{(c)}$$
$$+ B(|f(M)|) \sum_i \frac{|f(M)|}{|[f(M)]_i|} R_{ai}^2 \sum_j \sum_t Q_{bj}^{(a)} Q_{dt}^{(a)}$$
$$= A(|M|) + B(|M|)|M| \sum_i \frac{R_{ai}^2}{|[f(M)]_i|}.$$

which, since $R_{ai}$ is either 0 or $\frac{|[f(M)]_i|}{|M_a|}$, is

$$= A(|M|) + B(|M|)|M| \sum_{i: R_{ai} \neq 0} \frac{R_{ai}}{|M_a|}$$
$$= A(|M|) + B(|M|) \frac{|M|}{|M_a|} = g_M^{(k,m)}(\partial_{ab}, \partial_{cd}).$$

Finally, for the case $a = c, b = d$ equation (19) becomes

$$A(|M|) + B(|M|) \frac{|M|}{|M_a|} + C(|M|)|M| \sum_{i=1}^l \sum_{j=1}^n \frac{(R_{ai} Q_{bj}^{(a)})^2}{[f(M)]_{ij}}.$$

By (4) we have that $R_{ai} Q_{bj}^{(a)}$ is either $[f(M)]_{ij}/M_{ab}$ or 0. It follows then that equation (19) equals

$$= A(|M|) + B(|M|) \frac{|M|}{|M|_a}$$
$$+ C(|M|)|M| \sum_{i: R_{ai} \neq 0} \sum_{j: Q_{bj}^{(a)} \neq 0} \frac{R_{ai} Q_{bj}^{(a)}}{M_{ab}} = g_M^{(k,m)}(\partial_{ab}, \partial_{cd}).$$

We have shown that for any $M$ and $f \in \mathfrak{F}_{k,m}^{l,n}$

$$g_M^{(k,m)}(\partial_{ab}, \partial_{cd}) = g_{f(M)}^{(l,n)}(f_*\partial_{ab}, f_*\partial_{cd})$$

for each pair of tangent vectors in the basis $\{\partial_{ij} : i = 1, \ldots, k \ j = 1, \ldots, m\}$ and hence

$$f : \left(\mathbb{R}_+^{(k,m)}, g^{(k,m)}\right) \to \left(\mathbb{R}_+^{(l,n)}, g^{(l,n)}\right)$$

is an isometry. □

### 5.3 Normalized Conditional Models

A stronger statement can be said in the case of normalized conditional models. In this case, it turns out that the choices of $A$ and $B$ are immaterial and equation (12) reduces to the product Fisher information, scaled by a constant that represents the choice of $C$.

**Corollary 1.** *In the case of the manifold of normalized conditional models, equation (12) in theorem 1 reduces to the product Fisher information metric up to a multiplicative constant.*

*Proof.* For $u, v \in T_M \mathbb{P}_{m-1}^k$ expressed in the coordinates of the embedding tangent space $T_M \mathbb{R}_+^{k \times m}$

$$u = \sum_{ij} u_{ij} \partial_{ij} \quad v = \sum_{ij} v_{ij} \partial_{ij}$$

we have

$$g_M^{(k,m)}(u,v) = \left(\sum_{ij} u_{ij}\right)\left(\sum_{ij} v_{ij}\right) A(|M|)$$

$$+ \sum_i \left(\sum_j u_{ij}\right)\left(\sum_j v_{ij}\right) \frac{|M|}{|M_i|} B(|M|)$$

$$+ \sum_{ij} u_{ij} v_{ij} \frac{|M| C(|M|)}{M_{ij}} = kC(k) \sum_{ij} \frac{u_{ij} v_{ij}}{M_{ij}}$$

since $|M| = k$ and $\forall v \in T_M \mathbb{P}_{m-1}^k$ we have $\sum_j v_{ij} = 0$ for all $i$. We see that the choice of $A$ and $B$ is immaterial and the resulting metric is precisely the product Fisher information metric up to a multiplicative constant $kC(k)$, that corresponds to the choice of $C$. □

## 6 A Geometric Interpretation of Logistic Regression and AdaBoost

The conditional $I$-divergence between two positive conditional models $p(y|x), q(y|x)$ with respect to a probability measure $r$ over $\mathcal{X}$ is

$$D_r(p, q) = \qquad (20)$$
$$\sum_x r(x) \sum_y \left( p(y|x) \log \frac{p(y|x)}{q(y|x)} - p(y|x) + q(y|x) \right).$$

Assuming $\epsilon = q - p \to 0$ we may approximate $D_r(p, q) = D_r(p, p + \epsilon)$ by a second order Taylor approximation around $\epsilon = 0$

$$D_r(p,q) \approx D_r(p,p) + \sum_{xy} \frac{\partial D(p, p+\epsilon)}{\partial \epsilon(y,x)}\bigg|_{\epsilon=0} \epsilon(y,x)$$
$$+ \frac{1}{2} \sum_{x_1 y_1} \sum_{x_2 y_2} \frac{\partial^2 D(p, p+\epsilon)}{\partial \epsilon(y_1, x_1) \partial \epsilon(y_2, x_2)}\bigg|_{\epsilon=0} \epsilon(y_1, x_1) \epsilon(y_2, x_2).$$

The first order terms

$$\frac{\partial D_r(p, p+\epsilon)}{\partial \epsilon(y_1, x_1)} = r(x_1) \left(1 - \frac{p(y_1|x_1)}{p(y_1|x_1) + \epsilon(y_1, x_1)}\right)$$

zero out for $\epsilon = 0$. The second order terms

$$\frac{\partial^2 D_r(p, p+\epsilon)}{\partial \epsilon(y_1, x_1) \partial \epsilon(y_2, x_2)} = \frac{\delta_{y_1 y_2} \delta_{x_1 x_2} r(x_1) p(y_1|x_1)}{(p(y_1|x_1) + \epsilon(y_1, x_1))^2}$$

at $\epsilon = 0$ are $\delta_{y_1 y_2} \delta_{x_1 x_2} \frac{r(x_1)}{p(y_1|x_1)}$. Substituting these expressions in the Taylor approximation gives

$$D_r(p, p+\epsilon) \approx \frac{1}{2} \sum_{xy} \frac{r(x) \epsilon^2(y,x)}{p(y|x)} = \frac{1}{2} \sum_{xy} \frac{(r(x)\epsilon(y,x))^2}{r(x) p(y|x)}$$

which is the squared length of $r(x)\epsilon(y, x) \in T_{r(x)p(y|x)} \mathbb{R}_+^{k \times m}$ under the metric (12) for the choices $A(|M|) = B(|M|) = 0$ and $C(|M|) = 1/(2|M|)$.

Lebanon and Lafferty (2001) have shown that maximum likelihood for logistic regression and minimum exponential loss for AdaBoost may be cast as minimization of conditional $I$-divergence $D_r$ subject to linear constraints. In both cases $r$ is the empirical distribution $r(x) = \frac{1}{N} \sum_{i=1}^N \delta_{x,x_i}$ over the training set $\{(x_i, y_i)\}_{i=1}^N$.

The $I$ divergence $D_r(p, q)$ which both logistic regression and AdaBoost minimize is then approximately the geodesic distance between the conditional models $r(x)p(y|x)$ and $r(x)q(y|x)$ under a metric (12) with the above choices of $A, B, C$. The fact that the models $r(x)p(y|x)$ and $r(x)q(y|x)$ are not strictly positive is not problematic, since by the continuity of the metric, theorems 1 and 2 pertaining to $\mathbb{R}_+^{k \times m}$ apply also to its closure $\overline{\mathbb{R}_+^{k \times m}}$ - the set of all non-negative conditional models.

## 7 Discussion

We formulated and proved an axiomatic characterization of a family of metrics, the simplest of which is the Fisher information metric in the conditional setting for both normalized and non-normalized models. This result is a strict generalization of Campbell and Čencov's theorems. For the case $k = 1$, Theorems 1 and 2 reduce to the theorem of Campbell (1986) and corollary 1 reduces to Lemma 11.3 of Čencov (1982).

Using the characterization theorem and the result of Lebanon and Lafferty (2001) we give for the first time, a differential geometric interpretation of logistic regression and AdaBoost whose metric is characterized by natural invariance properties. It is interesting to explore directions suggested by alternative choices of $A, B$ and $C$. Another interesting prospect is exploring conditional modeling, where the unbounded $I$ divergence is replaced with the more robust geodesic distance.